\documentclass[conference]{IEEEtran}
\IEEEoverridecommandlockouts

\usepackage{cite}
\usepackage{amsmath,amssymb,amsfonts}
\usepackage{graphicx}
\usepackage{textcomp}
\usepackage{xcolor}
\usepackage{booktabs}
\usepackage{makecell}
\usepackage{array}
\usepackage{tabularx}
\usepackage{url}
\usepackage[table]{xcolor}
\usepackage{url}
\definecolor{BestBlue}{HTML}{AED9EB}
\definecolor{SecondBlue}{HTML}{9BE0B7}
\definecolor{ThirdBlue}{HTML}{B7B8BD}

\newcommand{\best}[1]{\cellcolor{BestBlue}{\bfseries #1}}
\newcommand{\second}[1]{\cellcolor{SecondBlue}#1}
\newcommand{\third}[1]{\cellcolor{ThirdBlue}#1}

\DeclareRobustCommand{\bestlegend}{\colorbox{BestBlue}{\strut\textbf{best}}}
\DeclareRobustCommand{\secondlegend}{\colorbox{SecondBlue}{\strut\textbf{second-best}}}
\DeclareRobustCommand{\thirdlegend}{\colorbox{ThirdBlue}{\strut third-best}}

\def\BibTeX{{\rm B\kern-.05em{\sc i\kern-.025em b}\kern-.08em
    T\kern-.1667em\lower.7ex\hbox{E}\kern-.125emX}}

\begin{document}
\vspace{-1em}
\title{Bayesian Networks with Latent Time Embedding for Stage-Aware Causal Modeling of Alzheimer's Disease Progression}

\author{\IEEEauthorblockN{Nguyen Linh Dan Le}
\IEEEauthorblockA{\textit{Department of Biomedical Engineering} \\
\textit{The University of Melbourne, Melbourne, Australia }\\
dan.le@iee.org}
}


\maketitle

\begin{abstract}
Alzheimer’s disease (AD) progression is often described through the amyloid–tau–neurodegeneration, or AT(N), cascade. However, most longitudinal models represent this cascade either as a fixed sequence of biomarkers or as a black-box forecasting task. This makes it difficult to determine when biologically guided biomarker relationships influence future regional pathology. In this study, we introduce Bayesian Networks with Latent Time Embedding (BN-LTE), a Bayesian structural framework for stage-aware modeling of AD progression. BN-LTE estimates disease pseudotime from baseline biomarker profiles and constrains directed dependencies according to biologically plausible AT(N) ordering. Posterior spline-varying structural equations are then used to link initial multimodal measurements with future annualized regional tau-PET change. Across repeated subject-disjoint evaluations using ADNI data, BN-LTE shows strong spatial reconstruction of tau progression compared with the included forecasting baselines. Beyond spatial reconstruction, BN-LTE recovers posterior stage-varying AT(N)-constrained effects and identifies a mid-pseudotime window of amyloid sensitivity. This window is supported by model-implied g-formula contrasts, root-adjusted AIPW, mechanism-sensitive ablations, and robustness analyses across spline and prior specifications. Overall, these findings position BN-LTE as a Bayesian structural framework for forecasting tau progression while examining stage-dependent AT(N)-cascade mechanisms in observational longitudinal neuroimaging data. Our code is available at https://github.com/danleneurocom/BN-LTE.
\end{abstract}


\section{Introduction}
Alzheimer's disease (AD) unfolds over a long preclinical period in which amyloid deposition, tau aggregation, neurodegeneration, and cognitive decline emerge at different rates. The amyloid-tau-neurodegeneration AT(N) framework organizes these processes into a biological cascade \cite{Jack_2018}, but it does not specify when one biomarker becomes an active driver of another. A central modeling question is therefore whether AT(N) should be treated as a fixed sequence or as a stage-dependent system whose directed relationships vary across latent disease time. Resolving this question requires models that align individuals by disease stage and connect initial biomarker profiles to later regional pathology. Crucially, disease stage must be estimated without using future outcomes.

Tau-PET provides a natural target for studying this problem. Regional tau burden is strongly associated with clinical impairment and follows the anatomical staging pattern described by Braak and Braak \cite{Braak_1991}. However, future tau accumulation is not determined by existing regional tau burden alone. It also depends on amyloid context, tissue vulnerability, and network organization. An effective progression model should, therefore, not only reconstruct the spatial pattern of tau change but also clarify which baseline biomarker relationships contribute to that change at different disease stages.

Current modeling approaches capture only part of this objective. Biophysical propagation models encode plausible mechanisms of spread, but they often assume global or stationary dynamics. Machine-learning and deep forecasting models can learn flexible longitudinal mappings, but their predictions are difficult to interpret as biologically ordered mechanisms. Disease-staging models align individuals along latent progression axes, yet they typically do not estimate posterior-directed effects linking baseline biomarker profiles to later regional tau change. These limitations motivate a framework that combines spatial forecasting with stage-aware structural interpretation.

We propose Bayesian Networks with Latent Time Embedding (BN-LTE), a Bayesian structural framework for stage-aware modeling of AD progression. BN-LTE organizes multimodal longitudinal observations into index-visit and subsequent-visit pairs. It estimates disease pseudotime from baseline biomarker profiles within the training split, constrains directed dependencies according to biologically admissible AT(N) ordering, and fits posterior spline-varying structural equations to model future annualized regional tau-PET change. In this formulation, the AT(N) cascade is treated as a testable, stage-varying structural system rather than as a fixed sequence or a black-box predictor.

This study establishes BN-LTE as a stage-aware structural model for investigating AD progression using longitudinal neuroimaging data. Our contributions are threefold. i) We introduce an AT(N)-constrained Bayesian transition network in which posterior spline-varying parent effects characterize how biomarker relationships change across latent disease time. ii) We evaluate BN-LTE on longitudinal ADNI\footnote{Alzheimer's Disease Neuroimaging Initiative: \url{https://adni.loni.usc.edu}.} tau-PET data, include an OASIS3\footnote{Open Access Series of Imaging Studies: \url{https://www.oasis-brains.org}.} cross-cohort robustness analysis using shared variables, and compare the model with related state-of-the-art baselines. iii) Empirically, BN-LTE achieves strong spatial reconstruction of tau progression among the included baselines and identifies a mid-pseudotime amyloid-sensitive window for future tau accumulation. This finding is supported by posterior g-formula contrasts, root-adjusted AIPW, mechanism-sensitive ablations, and spline/prior robustness analyses, suggesting that BN-LTE can generate anatomically and biologically interpretable hypotheses about stage-dependent AT(N) mechanisms.

\begin{figure*}[t]
    \centering
    \includegraphics[width=\linewidth]{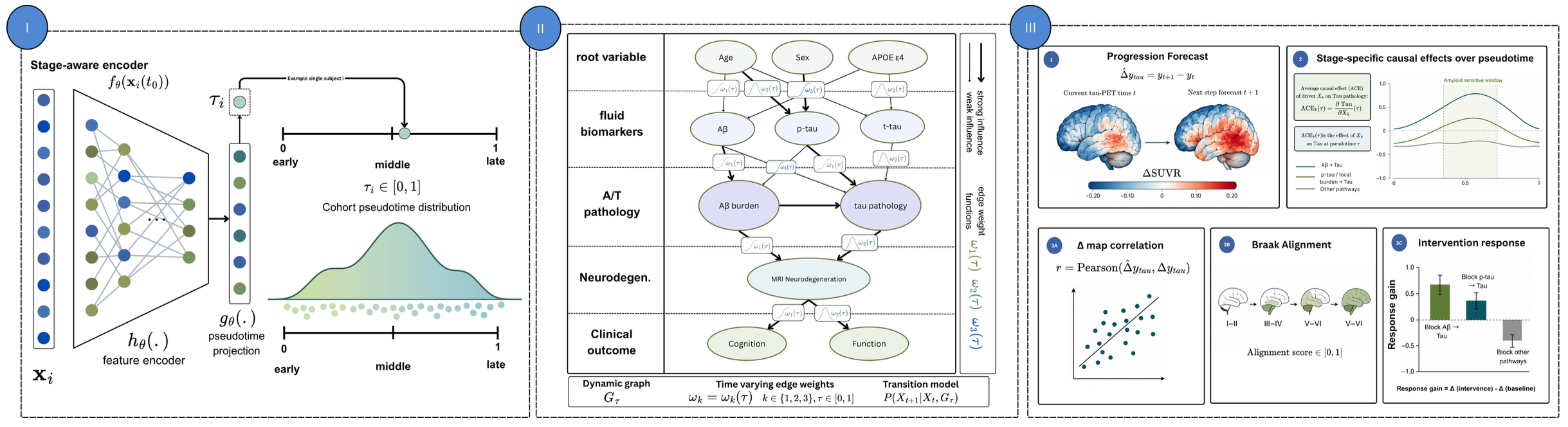}
    \caption{Overview of BN-LTE.
    Multimodal longitudinal observations are organized into index-visit and
    subsequent-visit pairs, where initial biomarker profiles define candidate
    causes and annualized regional tau-PET change defines the prediction target.
    A disease-time embedding maps initial biomarker profiles onto a pseudotime
    coordinate $\tau_i\in[0,1]$. BN-LTE represents the AT(N) cascade as an
    acyclic, biologically constrained Bayesian transition network, with directed
    dependencies flowing from root variables and fluid biomarkers through A/T
    pathology, neurodegeneration, and clinical outcomes. Edge functions are
    modeled as posterior spline-varying effects $b_{j\ell}(\tau)$, allowing
    biomarker relationships to strengthen or weaken across latent disease time.
    The fitted model predicts regional tau-PET progression in independent test
    subjects and supports stage-specific model-implied intervention analyses.}
\end{figure*}

\section{Related Work}
\label{sec:related_work}

\paragraph{Biophysical propagation  and staging models.}
A major line of AD progression modeling explains pathology as a spatial
spreading process over brain structure.  The network
diffusion model (NDM) was introduced to model dementia progression
\cite{Raj_2012} and was later used to predict longitudinal atrophy and
metabolism patterns in Alzheimer's disease \cite{Raj_2015}. Epidemic
spreading models extend the same network view by treating misfolded proteins
as transmissible agents over brain networks \cite{Iturria_Medina_2014}. A
related Bayesian physics-based formulation estimates uncertainty in tau
propagation parameters from tau-PET data \cite{Schafer_2021}. These approaches
are useful mechanistic models because they encode a spatial spreading
hypothesis, but their dynamics are usually specified by a fixed propagation
law. In contrast, event-based staging models estimate an ordering of biomarker
abnormality events \cite{Fonteijn_2012}. They provide a probabilistic disease
stage, but do not by themselves estimate directed, stage-varying effects from
baseline multimodal biomarkers to future regional tau rates.

\paragraph{Machine-learning prediction of future tau pathology.}
At the other end of the spectrum, supervised machine-learning models treat AD
progression as a flexible prediction problem. Giorgio et al.
developed a robust, interpretable prognostic index from multimodal biological
data to stratify subjects and predict future pathological tau accumulation
\cite{Giorgio_2022}. Rathore et al. used tau-PET radiomics together with
clinical, demographic, amyloid, and genetic variables to predict regional
future tau accumulation and to separate stable/slow from fast-progressing
regions \cite{Rathore_2024}. Karlsson et al. focused on predicting tau-PET in
Alzheimer's disease from plasma, MRI, and clinical variables, with code
released for their machine-learning pipeline \cite{Karlsson_2025}. These
studies are the closest forecasting comparators to our endpoint as they are supervised predictors of tau pathology, whereas BN-LTE is designed to
retain competitive prediction while exposing a stage-dependent structural
model of biomarker effects.

\paragraph{Multimodal deep learning and graph-learning comparators.}
Recent deep and graph-learning studies broaden Alzheimer's disease prediction
beyond conventional tabular regression by fusing multimodal biomarkers,
patient-similarity graphs, longitudinal structure, and brain-connectivity
topology. Jasodanand et al. used multimodal fusion to predict amyloid and tau
PET positivity across cohorts \cite{Jasodanand_2025}, while Ozdemir et al.
combined graph convolutional patient representations with tensor-algebra
temporal modeling for early AD prediction \cite{Ozdemir_2024}. Tekkesinoglu
and Pudas framed ADNI diagnosis as population-graph classification with
post-hoc explanations \cite{Tekkesinoglu_2024}; Lauber et al. instead used
graphical-lasso tau dependency networks to study how amyloid burden reshapes
tau-propagation topology \cite{Lauber_2024}. HiBrain further emphasizes
interpretable multimodal brain graphs, organizing functional and structural
connectivity through node-, graph-, and stage-level prototypes for stage-aware
neurodegenerative disease classification \cite{Ren_2026_HiBrain}. Together,
these works demonstrate the value of multimodal and graph-structured
representations, but their primary endpoints remain biomarker positivity,
diagnosis/progression classification, or population-level topology analysis
rather than continuous regional tau-rate forecasting.

\paragraph{Position of BN-LTE.}
BN-LTE is designed to connect these lines of work by retaining the spatial
forecasting objective of progression models, but constrains the explanation
through AT(N)-compatible directionality. We leverage latent disease time to make
biomarker effects stage-dependent, and uses posterior spline-varying structural
equations to quantify uncertainty in those effects. This allows BN-LTE to move
beyond fixed propagation laws and black-box forecasting. The resulting model
treats the AT(N) cascade as a computationally testable structural system whose
directed relationships can change across disease stage.

\section{Methodology}
\label{sec:methodology}

BN-LTE models AD progression as a Bayesian structural transition from an index
visit to later regional tau-PET accumulation. For each longitudinal pair $i$,
let $X_i^0\in\mathbb{R}^{p}$ denote multimodal measurements at the index
visit, and let $Y_i^0,Y_i^1\in\mathbb{R}^{q}$ denote regional tau-PET
measurements at the index and later visits. The main experiment uses $q=68$
Desikan--Killiany cortical regions, and the annualized regional tau
accumulation rate is $R_{ij}=(Y_{ij}^{1}-Y_{ij}^{0})/\Delta_i$, where
$\Delta_i$ is elapsed time in years. All splits are performed by participant.
All preprocessing, disease time estimation, parent selection, and tuning are
learned within the training split and then applied unchanged to validation,
test, and external cohort analyses.

BN-LTE has three linked components. First, it assigns each subject pair to a
latent disease time coordinate. Second, it restricts candidate dependencies
using biologically admissible AT(N) ordering. Third, it estimates posterior
structural equations whose effects vary smoothly over disease time.

\subsection{Latent Disease Time}

BN-LTE assigns each pair a scalar disease stage $Z_i\in[0,1]$ from index visit
biomarkers. Let $M_i\subseteq X_i^0$ denote the pseudotime feature set and
$\widetilde M_i$ its normalized representation. The primary disease time
coordinate is obtained from the leading component of the training matrix:
\begin{equation}
\widetilde M_{\mathrm{train}}=U\Sigma V^\top,\qquad
s_i=\widetilde M_i^\top V_{\cdot 1},\qquad
Z_i=\mathcal Q_{\mathrm{train}}(s_i),
\label{eq:latent_time}
\end{equation}
Here, $\mathcal Q_{\mathrm{train}}$ maps scores to $[0,1]$ using training
percentiles, after orienting the component so that larger values indicate
greater pathological burden. The primary pseudotime excludes regional tau-PET
targets and pTau217, separating disease stage estimation from the tau
accumulation endpoints.

To test whether the mechanistic findings depend on this linear staging axis,
we also evaluate an AT(N) monotone latent time model. It estimates a scalar
$\tau_i\in[0,1]$ and smooth trajectories $f_m(\tau)$ for groups of biomarkers:
\begin{equation}
\begin{aligned}
\min_{\{\tau_i\},\{f_m\}} \quad
&\sum_{(i,m)\in\Omega}
\bigl(\widetilde M_{im}-f_m(\tau_i)\bigr)^2  \\
&+\lambda_{\mathrm{mono}}\mathcal L_{\mathrm{mono}}
+\lambda_{\mathrm{ATN}}\mathcal L_{\mathrm{ATN}}
+\lambda_{\mathrm{smooth}}\mathcal L_{\mathrm{smooth}} .
\end{aligned}
\label{eq:atn_monotone}
\end{equation}
The monotonicity term enforces biologically expected directions of change, and
the AT(N) ordering term encourages amyloid, tau, neurodegeneration, and
clinical changes to emerge in a biologically plausible order. This variant is
used as a mechanistic audit of the disease time construction. 

\subsection{AT(N) Constrained Transition Graph}

Given a disease time coordinate, BN-LTE next defines which variables can act as
parents of later tau accumulation. Variables are assigned to ordered biological
layers:
$
\mathrm{root}\prec\mathrm{fluid}\prec\mathrm{pathology}\prec
\mathrm{neurodegeneration}\prec\mathrm{clinical}.
$
Root variables include age, sex, education, and APOE $\varepsilon4$ dosage.
Clinical variables are treated as sinks. For each tau-PET region $j$, the
parent set $\operatorname{pa}(j)$ is selected within the training split using
AT(N) directionality, measurement coverage, and association with $R_{ij}$.
Regional self history is modeled separately through $Y_{ij}^0$.

The resulting transition network factorizes the regional tau accumulation
model as
\begin{equation}
p(R_i\mid X_i^0,Y_i^0,Z_i)=
\prod_{j=1}^{q}
p\!\left(
R_{ij}\mid Y_{ij}^{0},
X_{i,\operatorname{pa}(j)}^0,
Z_i
\right).
\label{eq:factorization}
\end{equation}
This graph encodes the temporal and biological orientation of BN-LTE: causes
are measured at the index visit, while outcomes are later regional tau
accumulation rates.

\begin{table*}[t]
\centering
\caption{Repeated subject-disjoint ADNI/OASIS3 all-68-region tau-progression benchmarks.
Values are mean$\pm$SD across splits; \bestlegend, \secondlegend, and
\thirdlegend denote top distinct means within each dataset block.}
\label{tab:repeated_benchmark}
\scriptsize
\setlength{\tabcolsep}{2.0pt}
\renewcommand{\arraystretch}{1.08}
\resizebox{\textwidth}{!}{%
\begin{tabular}{@{}ll*{6}{c}*{6}{c}@{}}
\toprule
Model & Type & \multicolumn{6}{c}{ADNI} & \multicolumn{6}{c}{OASIS3} \\
\cmidrule(lr){3-8}\cmidrule(lr){9-14}
& & MAE$\downarrow$ & $\rho_\Delta\uparrow$ & Cos.$\uparrow$ & Top-3$\uparrow$ & Braak $\rho\uparrow$ & B-MAE$\downarrow$ & MAE$\downarrow$ & $\rho_\Delta\uparrow$ & Cos.$\uparrow$ & Top-3$\uparrow$ & Braak $\rho\uparrow$ & B-MAE$\downarrow$ \\
\midrule
Raj et al. {\scriptsize\cite{Raj_2012}} & Biophysical & 1.05$\pm$0.26 & 0.77$\pm$0.06 & 0.91$\pm$0.03 & \third{0.94$\pm$0.04} & 0.64$\pm$0.48 & 0.94$\pm$0.40 & 3.05$\pm$1.56 & 0.02$\pm$0.11 & 0.05$\pm$0.44 & 0.22$\pm$0.23 & \best{0.32$\pm$0.58} & 3.39$\pm$1.78 \\
Iturria-Medina et al. {\scriptsize\cite{Iturria_Medina_2014}} & Biophysical & 1.04$\pm$0.21 & 0.76$\pm$0.06 & 0.92$\pm$0.03 & \best{0.96$\pm$0.02} & 0.64$\pm$0.36 & 1.01$\pm$0.44 & 3.05$\pm$1.56 & 0.02$\pm$0.11 & 0.05$\pm$0.44 & 0.22$\pm$0.23 & \best{0.32$\pm$0.58} & 3.39$\pm$1.78 \\
Schafer et al. {\scriptsize\cite{Schafer_2021}} & Bayesian/Biophysical & 1.11$\pm$0.28 & 0.76$\pm$0.07 & 0.89$\pm$0.05 & \third{0.94$\pm$0.03} & \second{0.72$\pm$0.52} & 0.88$\pm$0.41 & 3.05$\pm$1.56 & 0.02$\pm$0.11 & 0.05$\pm$0.44 & 0.22$\pm$0.23 & \best{0.32$\pm$0.58} & 3.39$\pm$1.78 \\
\midrule
Giorgio et al. {\scriptsize\cite{Giorgio_2022}} & Machine Learning & 1.17$\pm$0.47 & 0.64$\pm$0.29 & \best{0.98$\pm$0.02} & 0.90$\pm$0.09 & 0.60$\pm$0.35 & 1.02$\pm$0.65 & \second{1.39$\pm$0.18} & 0.08$\pm$0.43 & 0.10$\pm$0.45 & 0.62$\pm$0.35 & 0.07$\pm$0.64 & \second{1.46$\pm$0.19} \\
Karlsson et al. {\scriptsize\cite{Karlsson_2025}} & Machine Learning & \third{0.90$\pm$0.57} & 0.70$\pm$0.23 & \best{0.98$\pm$0.01} & 0.92$\pm$0.11 & \best{0.73$\pm$0.31} & \second{0.78$\pm$0.67} & 1.64$\pm$0.45 & 0.22$\pm$0.54 & \third{0.25$\pm$0.26} & \second{0.67$\pm$0.38} & -0.13$\pm$0.70 & 1.71$\pm$0.43 \\
Rathore et al. {\scriptsize\cite{Rathore_2024}} & Machine Learning & 1.24$\pm$0.23 & 0.49$\pm$0.02 & \second{0.97$\pm$0.00} & 0.89$\pm$0.09 & 0.33$\pm$0.12 & 1.06$\pm$0.24 & 1.85$\pm$0.42 & \best{0.40$\pm$0.46} & 0.16$\pm$0.10 & 0.63$\pm$0.28 & 0.07$\pm$0.81 & 1.90$\pm$0.47 \\
\midrule
Tabarestani et al. {\scriptsize\cite{Tabarestani_2020}} & Deep learning & 4.52$\pm$0.77 & 0.09$\pm$0.27 & 0.77$\pm$0.21 & 0.73$\pm$0.10 & 0.27$\pm$0.46 & 2.89$\pm$0.76 & 8.92$\pm$2.74 & 0.16$\pm$0.43 & \second{0.29$\pm$0.39} & \third{0.66$\pm$0.10} & 0.07$\pm$0.64 & 5.72$\pm$2.61 \\
Nguyen et al. {\scriptsize\cite{Nguyen_2020}} & Deep learning & 2.51$\pm$0.74 & 0.30$\pm$0.33 & 0.91$\pm$0.04 & 0.86$\pm$0.02 & 0.27$\pm$0.58 & 1.91$\pm$0.71 & 5.92$\pm$0.62 & 0.21$\pm$0.30 & -0.05$\pm$0.49 & \third{0.66$\pm$0.32} & -0.13$\pm$0.61 & 3.72$\pm$1.72 \\
Jasodanand et al. {\scriptsize\cite{Jasodanand_2025}} & Deep learning & 1.43$\pm$0.52 & 0.22$\pm$0.32 & 0.93$\pm$0.05 & 0.80$\pm$0.04 & 0.40$\pm$0.53 & 0.85$\pm$0.32 & 6.66$\pm$4.20 & -0.33$\pm$0.36 & 0.07$\pm$0.42 & 0.47$\pm$0.41 & -0.20$\pm$0.87 & 3.07$\pm$1.42 \\
\midrule
Ozdemir et al. {\scriptsize\cite{Ozdemir_2024}} & Graph learning & 1.05$\pm$0.35 & 0.57$\pm$0.16 & \best{0.98$\pm$0.00} & 0.92$\pm$0.01 & 0.53$\pm$0.23 & 0.92$\pm$0.35 & \best{1.35$\pm$0.19} & \second{0.27$\pm$0.52} & \best{0.31$\pm$0.40} & 0.61$\pm$0.27 & 0.07$\pm$0.81 & \best{1.45$\pm$0.19} \\
Tekkesinoglu et al. {\scriptsize\cite{Tekkesinoglu_2024}} & Graph learning & \best{0.78$\pm$0.07} & 0.62$\pm$0.12 & \best{0.98$\pm$0.00} & 0.89$\pm$0.06 & 0.53$\pm$0.23 & \best{0.62$\pm$0.25} & 1.81$\pm$0.33 & -0.02$\pm$0.16 & -0.06$\pm$0.25 & 0.51$\pm$0.32 & -0.33$\pm$0.12 & 1.91$\pm$0.34 \\
Ren et al. {\scriptsize\cite{Ren_2026_HiBrain}} & Graph Learning & 0.98$\pm$0.66 & 0.54$\pm$0.26 & \best{0.98$\pm$0.01} & 0.90$\pm$0.09 & \third{0.67$\pm$0.42} & \third{0.83$\pm$0.62} & \third{1.59$\pm$0.21} & 0.13$\pm$0.35 & 0.20$\pm$0.33 & \best{0.73$\pm$0.09} & 0.07$\pm$0.64 & \third{1.67$\pm$0.21} \\
\midrule
BN-LTE & Ours & \second{0.89$\pm$0.42} & \best{0.84$\pm$0.04} & \third{0.94$\pm$0.03} & \second{0.95$\pm$0.07} & \second{0.72$\pm$0.11} & 0.88$\pm$0.46 & 2.28$\pm$1.09 & \third{0.25$\pm$0.16} & 0.21$\pm$0.53 & 0.34$\pm$0.20 & 0.08$\pm$0.44 & 2.20$\pm$1.54 \\
BN-LTE + PCA-Z & Ours & 1.04$\pm$0.56 & \third{0.81$\pm$0.12} & 0.91$\pm$0.08 & 0.93$\pm$0.05 & 0.48$\pm$0.41 & 1.05$\pm$0.99 & 2.30$\pm$1.15 & \second{0.27$\pm$0.14} & 0.23$\pm$0.54 & 0.27$\pm$0.22 & \second{0.24$\pm$0.46} & 2.30$\pm$1.52 \\
BN-LTE + ATN-Z tau-free & Ours & 0.91$\pm$0.32 & \second{0.82$\pm$0.10} & 0.92$\pm$0.07 & 0.92$\pm$0.07 & 0.52$\pm$0.34 & \third{0.83$\pm$0.53} & 2.25$\pm$1.06 & \third{0.25$\pm$0.13} & 0.23$\pm$0.54 & 0.36$\pm$0.23 & \third{0.12$\pm$0.34} & 2.25$\pm$1.44 \\
\bottomrule
\end{tabular}%
}
\end{table*}

\subsection{Posterior Structural Equations over Disease Time}

BN-LTE then estimates how each admissible parent acts over disease time. For
each target region $j$, the regional accumulation rate is modeled as
\begin{equation}
\begin{aligned}
R_{ij}
&=a_j(Z_i)+c_j(Z_i)Y_{ij}^{0}
+b_j(Z_i)^\top X_{i,\operatorname{pa}(j)}^0
+\varepsilon_{ij}
\end{aligned}
\label{eq:scm}
\end{equation}
where $\varepsilon_{ij}\sim\mathcal N(0,\sigma_j^2)$. The function $a_j(\cdot)$ captures regional drift, $c_j(\cdot)$ captures
regional self history, and $b_j(\cdot)$ contains the parent effects for region
$j$. These effects are allowed to strengthen or weaken along disease time. Each
coefficient function is expanded in a cubic spline basis:
\begin{equation}
a_j(z)=\alpha_j^\top B(z),\qquad
c_j(z)=\kappa_j^\top B(z),\qquad
b_j(z)=\Theta_j B(z),
\label{eq:spline_effects}
\end{equation}
where $B(z)\in\mathbb R^{K_B}$ and $K_B=4$ in the primary model. Sensitivity
analyses over $K_B$ and prior scale evaluate whether the inferred mechanisms
depend on these choices.

Let $r_j$ be the training vector of regional tau accumulation rates and
$\Phi_j$ the spline expanded design matrix. BN-LTE uses conjugate Bayesian
regression:
\begin{equation}
\begin{aligned}
r_j &= \Phi_j\beta_j+\varepsilon_j,
\qquad \varepsilon_j\sim\mathcal N(0,\sigma_j^2I),\\
\beta_j\mid\sigma_j^2 &\sim \mathcal N(0,\sigma_j^2V_0),
\qquad
\sigma_j^2\sim\mathrm{InvGamma}(a_0,b_0).
\end{aligned}
\label{eq:posterior}
\end{equation}
This yields an analytic posterior for each regional structural equation.
Posterior samples quantify uncertainty in disease time effect curves, edge
activity, predictions, and intervention contrasts. For an admissible parent
$\ell$, edge activity is summarized as
\[
P_{j\ell}^{\mathrm{act}}
=
P\!\left(\sup_{z\in[0,1]} |b_{j\ell}(z)|>\eta
\mid \mathcal D_{\mathrm{train}}\right),
\qquad \eta=0.01 .
\]

\subsection{Forecasting and Intervention Contrasts}

The posterior structural equations produce both predictions and mechanistic
contrasts. For a new pair, BN-LTE estimates $Z_i$, computes the posterior mean
accumulation rate $\widehat R_i$, and predicts later tau burden by
$\widehat Y_i^1=Y_i^0+\Delta_i\widehat R_i$. For model-implied interventions,
selected parent variables $X_{iS}^0$ are replaced by reference values
$x_S^\star$ inside the fitted structural equations. The posterior contrast at
disease stage $z$ is
\begin{equation}
\Delta_S(z)=
\mathbb E_{\mathrm{post}}
\!\left[
R_j\{X_S^0=x_S^\star,Z=z\}
-
R_j\{X_S^0,Z=z\}
\right].
\label{eq:gformula}
\end{equation}
These contrasts quantify stage-dependent model-implied shifts in tau
accumulation under the learned AT(N)-constrained structural graph.

\subsection{Evaluation and Calibration}

Evaluation focuses on spatial prediction, mechanistic validity, and uncertainty
calibration. Spatial performance is measured by regional error, delta-map
agreement, top-progression recovery, and Braak alignment. Mechanistic validity
is assessed through stage-varying effects, biological admissibility, the
mid-amyloid window, and intervention contrasts. Observational support is
examined with AIPW and placebo diagnostics, while uncertainty is summarized using posterior intervals and residual calibration.

\section{Experiments and Analysis}
\label{sec:experiments}

\subsection{Datasets and Baselines}

We evaluated BN-LTE on 541 ADNI participants forming 796 baseline-to-follow-up pairs, with 88 baseline predictors and 68 cortical tau-PET targets. Subject-disjoint splitting produced 477 training pairs, 167 validation pairs, and 152 held-out test pairs for the default split, and repeated analyses used five subject-disjoint train/validation/test splits. We further assessed cross-cohort robustness on OASIS3 using the feature subset shared with ADNI and 26 longitudinal AV1451 tau-PET pairs. Because OASIS3 is substantially smaller and lacks several ADNI plasma biomarkers, we treat it as a supporting robustness cohort rather than a primary model-selection cohort.

We compared BN-LTE against the models reported in Table~I, spanning four baseline families. Biophysical and Bayesian biophysical propagation baselines include NDM~\cite{Raj_2012}, ESM~\cite{Iturria_Medina_2014}, and Bayesian NDM~\cite{Schafer_2021}. Machine-learning baselines include ML-Prognostic Index~\cite{Giorgio_2022}, Karlsson Tau-PET ML~\cite{Karlsson_2025}, and AdaBoost Tau-Rate~\cite{Rathore_2024}. Neural and multimodal deep-learning baselines include DeepMTL-MLP~\cite{Tabarestani_2020}, ResidualDeepEnsemble~\cite{Nguyen_2020}, and NComms2025 Fusion MLP~\cite{Jasodanand_2025}. Graph-learning comparators include DyEPAD Dynamic Graph~\cite{Ozdemir_2024}, GCN-XAI Population Graph~\cite{Tekkesinoglu_2024}, and HiBrain~\cite{Ren_2026_HiBrain}. We report BN-LTE, BN-LTE + PCA-Z, and BN-LTE + ATN-Z tau-free under the same repeated subject-disjoint protocol.

\subsection{Quantitative Results}
As shown in Table~\ref{tab:repeated_benchmark}, BN-LTE provides the most
balanced performance on the primary ADNI benchmark. Although flexible ML and
graph-learning baselines achieve strong pointwise errors, BN-LTE attains the
best delta-map Spearman ($\rho_\Delta=0.84$) and ranks second in
MAE, Top-3 progression capture, and Braak-order correlation. This indicates
that the model does not simply smooth regional tau burden, but preserves the
direction, spatial concentration, and anatomical organization of future tau
change. The advantage over NDM and ESM further suggests that fixed global
propagation laws are insufficient to capture stage-dependent, subject-specific
tau-rate variation.

The OASIS3 results should be interpreted in light of the cohort and measurement
setting. Compared with ADNI, the available OASIS3 subset is smaller, has fewer
matched longitudinal tau-PET pairs, and lacks several baseline biomarker
channels used by BN-LTE, making absolute-error metrics more sensitive to cohort
composition, missing-feature imputation, and tracer/acquisition heterogeneity.
Under these constraints, some adapted deep or graph baselines obtain lower raw
MAE, but BN-LTE variants still retain competitive rank-based progression signal,
including top-three $\rho_\Delta$ and Braak-ordering performance. Overall,
BN-LTE is best understood as a stage-aware structural progression model:
it combines competitive forecasting with interpretable AT(N)-constrained,
disease-time-varying baseline-to-future-rate effects.


\subsection{Pseudotime Robustness and AT(N)-Monotone Extension}

\begin{figure}[t]
    \centering
    \includegraphics[width=0.98\linewidth]{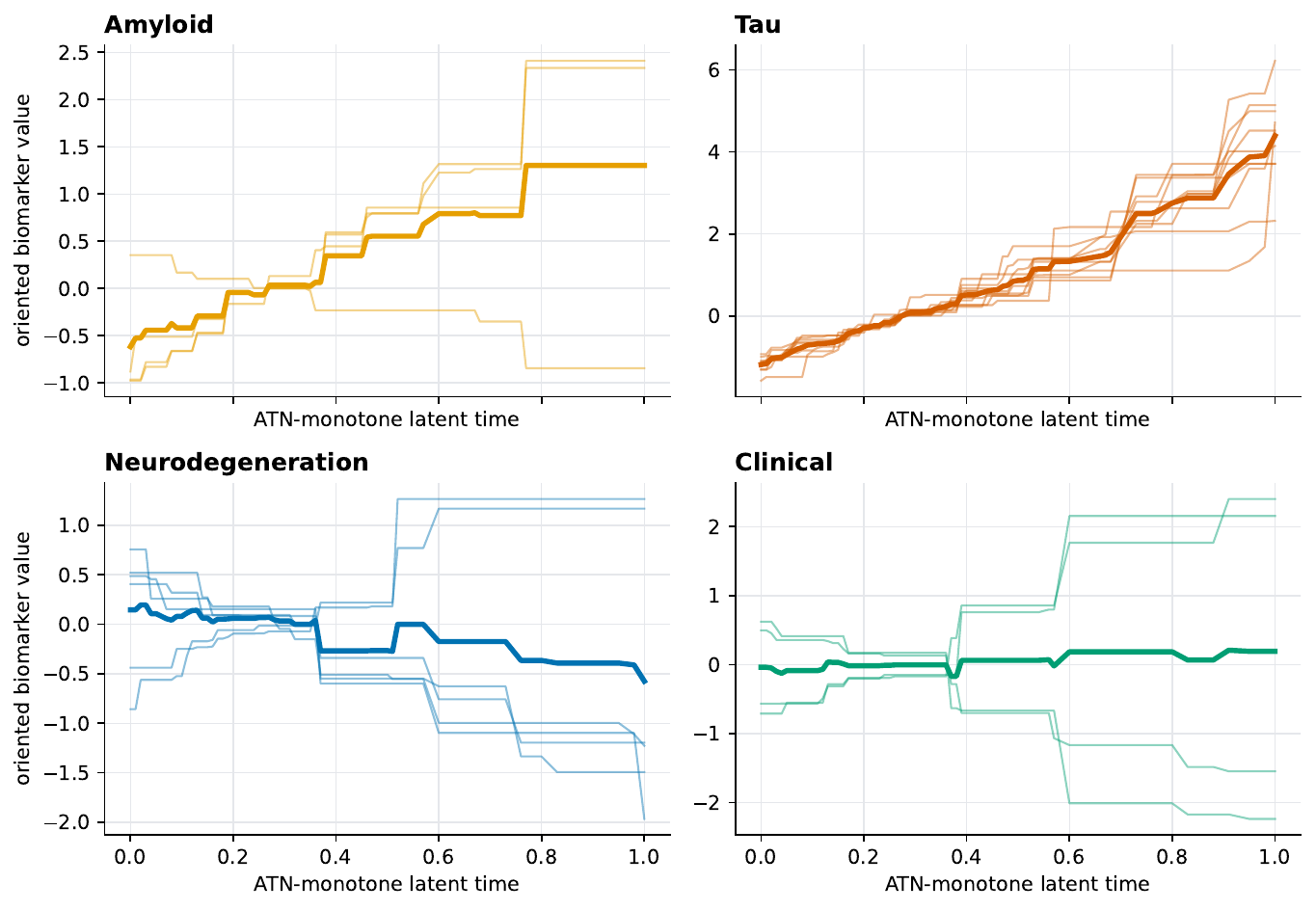}
    \caption{AT(N)-monotone latent-time trajectories.}
    \label{fig:atn_monotone_trajectories}
\end{figure}

The primary BN-LTE model estimates disease time from a tau-free biomarker
profile, so subject stage is not defined by the regional tau accumulation
endpoints. In this section, we examine whether the structural findings depend on this linear
disease-time coordinate. The AT(N)-monotone variant replaces the projection
with smooth biomarker trajectories constrained by biological directionality and
the expected AT(N) ordering. Figure~\ref{fig:atn_monotone_trajectories} shows that the constrained latent
time preserves an interpretable disease-stage structure. Amyloid and tau
burden increase along the latent axis, while neurodegeneration and clinical
measurements evolve coherently after orientation. This indicates that the
alternative staging model captures meaningful AD progression structure rather
than an arbitrary scalar ordering. Furthermore, the AT(N)-monotone variant preserves the
mid-pseudotime amyloid-sensitive window, whereas random or reversed pseudotime
weakens this signal. Thus, BN-LTE's central amyloid-window finding is not a
consequence of assigning any scalar stage variable. It remains visible under a
biologically constrained alternative to the primary disease-time embedding.

\subsection{Posterior Stage-Varying Effects}

\begin{figure}[t]
    \centering
    \includegraphics[width=0.98\linewidth]{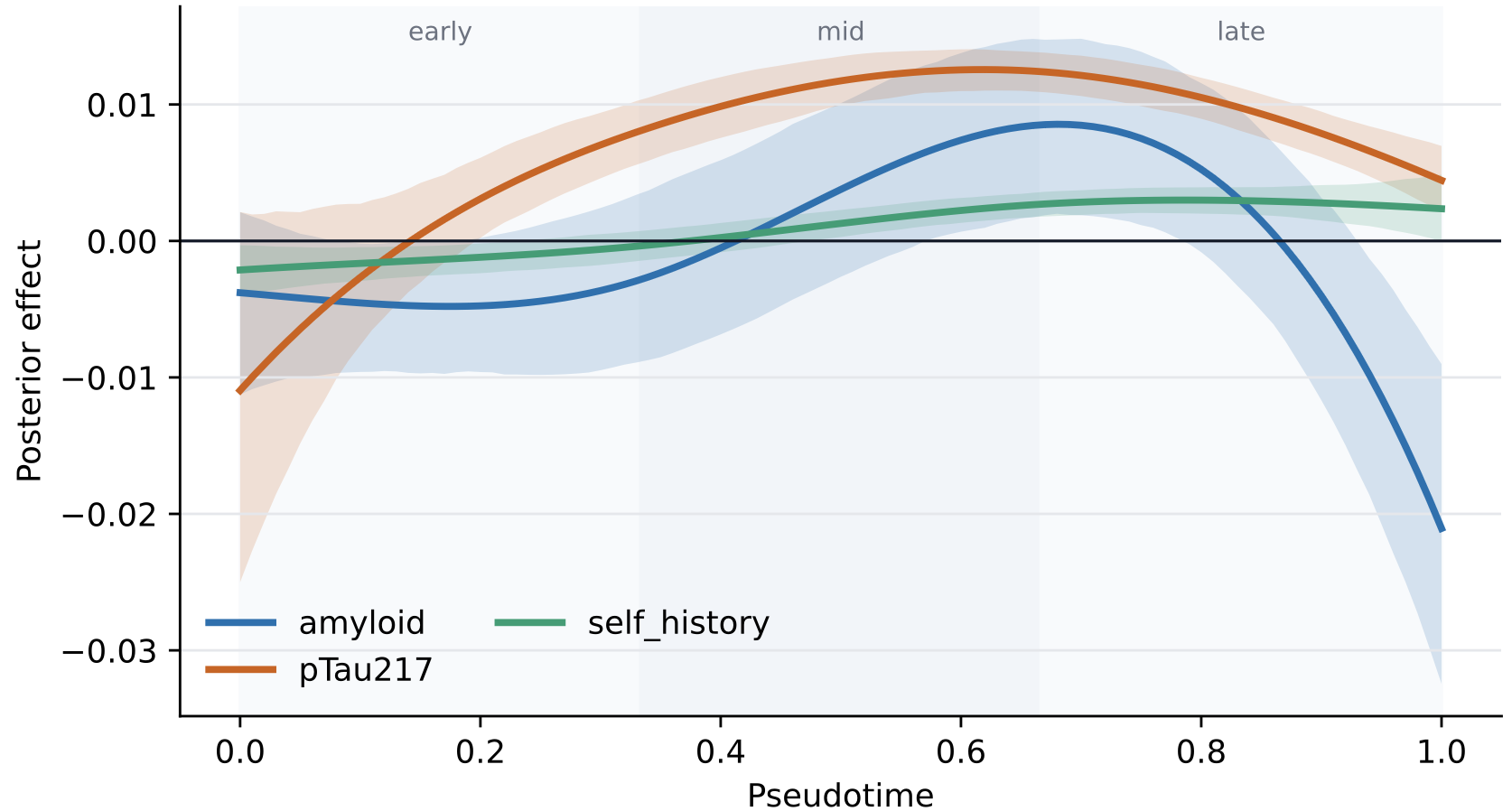}
    \caption{Posterior stage-varying structural effects in BN-LTE.}
    \label{fig:posterior_effect_curves}
\end{figure}

Figure~\ref{fig:posterior_effect_curves} shows that BN-LTE does not explain
tau progression through a single stationary biomarker effect. The pTau217 curve
becomes strongest through the middle of pseudotime, suggesting that soluble tau
pathology is most informative once disease has moved beyond the earliest stage.
Amyloid effects are more heterogeneous and carry wider uncertainty, especially
near the extremes of pseudotime, indicating that amyloid is not well captured
as a constant global driver. The self-history curve remains comparatively
stable, showing that BN-LTE separates regional persistence from upstream AT(N)
effects. This pattern motivates the g-formula analysis: the amyloid-sensitive
window is most clearly expressed through stage-conditioned structural
substitutions rather than through a single averaged coefficient curve.

\subsection{Mechanism-Sensitive Ablations}

\begin{table}[t]
\centering
\caption{Mechanism-sensitive ablations\protect\footnotemark}
\label{tab:mechanism_ablation}
\scriptsize
\setlength{\tabcolsep}{2.5pt}
\renewcommand{\arraystretch}{1.10}
\begin{tabular}{lccccc}
\toprule
Variant & Tau-rate MAE$\downarrow$ & $\rho_\Delta\uparrow$ & \makecell{Stage\\heterogeneity$\uparrow$} & \makecell{Mid-A$\beta$\\window$\uparrow$} & \makecell{Violation\\rate$\downarrow$} \\
\midrule
Full BN-LTE & \underline{0.0054} & 0.896 & 0.0070 & \textbf{0.0078} & \textbf{0.000} \\
Static BN, no latent time & 0.0058 & 0.881 & 0.0000 & -- & \textbf{0.000} \\
Random pseudotime & 0.0058 & 0.877 & \underline{0.0091} & 0.0002 & \textbf{0.000} \\
Reversed pseudotime & \underline{0.0054} & 0.896 & 0.0070 & -0.0002 & \textbf{0.000} \\
No spline-varying effects & 0.0058 & 0.881 & 0.0000 & -0.0024 & \textbf{0.000} \\
No AT(N) constraints & 0.0055 & 0.894 & 0.0071 & \textbf{0.0078} & 0.333 \\
Shuffled AT(N) layers & 0.0055 & 0.897 & \textbf{0.0126} & 0.0056 & \underline{0.005} \\
No self-history & 0.0055 & \underline{0.899} & 0.0066 & \underline{0.0076} & \textbf{0.000} \\
pTau217-free pseudotime & 0.0055 & \underline{0.899} & 0.0053 & -0.0004 & \textbf{0.000} \\
No amyloid parents & \textbf{0.0053} & \textbf{0.906} & 0.0019 & 0.0000 & \textbf{0.000} \\
No pTau217 parents & 0.0057 & 0.885 & 0.0079 & -0.0018 & \textbf{0.000} \\
\bottomrule
\end{tabular}
\end{table}
\footnotetext{Unlike Table~\ref{tab:repeated_benchmark}, Table~\ref{tab:mechanism_ablation} reports BN-LTE ablations on the native annualized tau-rate scale; therefore, the MAE values are not directly comparable.}

Table~\ref{tab:mechanism_ablation} shows that several ablations retain similar MAE and $\rho_\Delta$, suggesting that regularized prediction can recover part of the group-level tau-rate pattern. However, these variants fail in distinct mechanistic ways. Static and no-spline models remove disease-time variation and therefore cannot explain when biomarker effects emerge. Random and reversed pseudotime preserve competitive error but weaken the mid-amyloid window, indicating that disease-time alignment is important for the structural signal. Removing AT(N) constraints leaves prediction nearly unchanged but permits biologically inadmissible directions, while removing amyloid parents abolishes the amyloid-window response despite improving MAE. Thus, our framework is distinguished by maintaining competitive spatial reconstruction while exposing biologically admissible, disease-time-varying AT(N) relationships.
\subsection{Sensitivity to Spline, Prior, and Parent-Selection Choices}

\begin{table}[t]
\centering
\caption{Sensitivity of the final BN-LTE fit to spline basis size and prior/shrinkage scale}
\label{tab:spline_prior_sensitivity}
\scriptsize
\setlength{\tabcolsep}{4.2pt}
\renewcommand{\arraystretch}{1.10}
\begin{tabular}{lccc}
\toprule
Sensitivity & MAE range$\downarrow$ & $\rho_\Delta$ range$\uparrow$ & Mid-A$\beta$ window \\
\midrule
Spline basis $K_B=3$--$6$ & 0.0056--0.0059 & 0.892--0.907 & 0.0034--0.0052 \\
Prior scale $0.25$--$4\times$ & 0.0056--0.0061 & 0.895--0.915 & 0.0036--0.0061 \\
Parent cap 4--10 & 0.0064--0.0077 & 0.894--0.910 & -0.0008--0.0001 \\
\bottomrule
\end{tabular}
\end{table}

\begin{figure}[t]
    \centering
    \includegraphics[width=\linewidth]{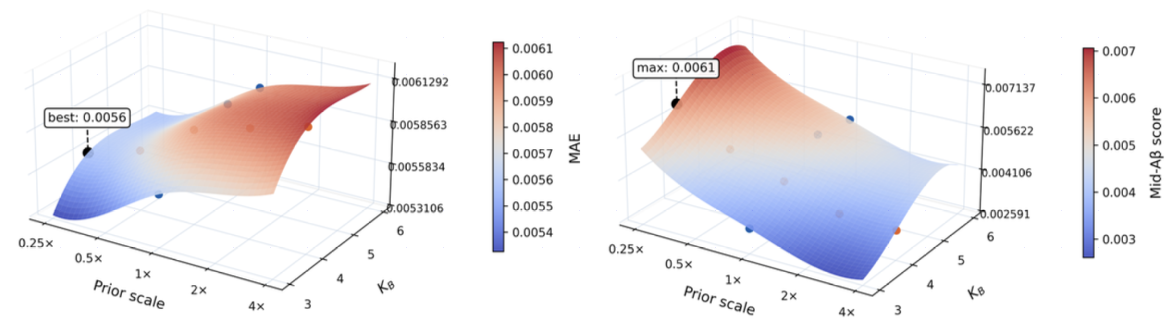}
\caption{Sensitivity of BN-LTE to spline basis size and prior/shrinkage scale.}
\label{fig:spline_prior_surface}    
\end{figure}

Table~\ref{tab:spline_prior_sensitivity} examines the stability of BN-LTE under alternative spline, prior, and parent-selection settings. Across spline basis sizes $K_B=3$--6 and prior/shrinkage scales from $0.25\times$ to $4\times$, prediction error and spatial topology vary only modestly, while the mid-A$\beta$ window remains positive. Visually demonstrated on Figure~\ref{fig:spline_prior_surface}, the response surface changes smoothly across the tested spline and prior settings, indicating that the amyloid-window signal is not tied to a single posterior smoothing configuration. Parent-cap sensitivity shows a different behavior. Although spatial agreement remains stable, the mid-A$\beta$ window weakens when the admissible parent set changes, reflecting the fact that parent selection controls which biological pathways can enter the structural equations. These results support the stability of the amyloid-window discovery under spline and prior variation, while showing that parent selection is the main source of mechanistic sensitivity.

\subsection{Spatial Progression Maps}

\begin{figure}[t]
    \centering
    \includegraphics[width=0.98\linewidth]{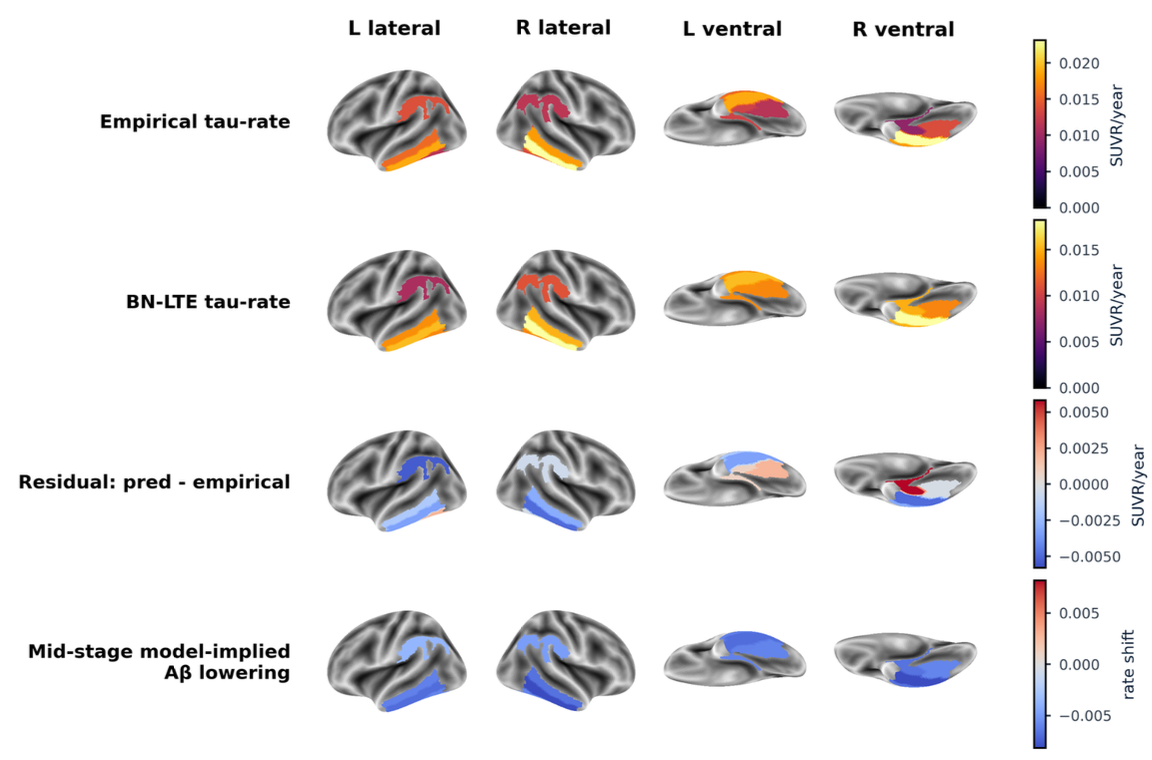}
    \caption{Spatial tau-progression reconstruction and model-implied amyloid response. Rows show empirical tau-rate, BN-LTE-predicted tau-rate, prediction residuals, and the mid-pseudotime effect of A$\beta$ lowering across lateral and ventral cortical views.}
    \label{fig:spatial_tau_progression}
\end{figure}

In Figure~\ref{fig:spatial_tau_progression} we compare the quantitative benchmark results of Table~\ref{tab:repeated_benchmark} to the anatomical distribution of predicted tau progression. The empirical tau-rate map shows that subsequent accumulation is concentrated in temporo-parietal and ventral temporal regions, consistent with the spatial pattern expected for AD-related tau spread . The corresponding BN-LTE prediction preserves this large-scale cortical topology, suggesting that the model captures structured regional progression, in contrast to solely reducing aggregate error. Subsequently, the residual map identifies localized deviations from this pattern, in addition to the intervention map provides a mechanistic readout by highlighting regions in which the fitted structural equations assign the strongest mid-pseudotime amyloid-lowering effect on future tau accumulation.

\subsection{Stage-Specific Intervention and Observational Causal Diagnostics}

\begin{table}[t]
\centering
\caption{Causal and intervention diagnostics. AIPW estimates observational amyloid associations; g-formula rows are model-implied substitutions inside BN-LTE}
\label{tab:causal_diagnostics}
\scriptsize
\setlength{\tabcolsep}{0.8pt}
\renewcommand{\arraystretch}{1.1}
\begin{tabular}{llcc}
\toprule
Analysis & Estimand & Estimate & Diagnostic \\
\midrule
AIPW root-only & mean tau-rate & 0.0135 & 95\% CI 0.0066--0.0203; ESS 527 \\
AIPW expanded & mean tau-rate & 0.0024 & 95\% CI -0.0103--0.0152; ESS 211 \\
Trimmed AIPW & root-only 0.05--0.95 & 0.0133 & 95\% CI 0.0061--0.0205 \\
Trimmed AIPW & root-only 0.10--0.90 & 0.0136 & 95\% CI 0.0069--0.0203 \\
Placebo refutation & permuted treatment & 0.0135 & $p=0.005$ \\
Stage AIPW & mid pseudotime & 0.0366 & 95\% CI 0.0051--0.0681; ESS 9.2 \\
Stage AIPW & late pseudotime & -- & 11 treated / 1 control \\
\midrule
$do(A\beta\ \mathrm{low})$ & early rate shift & -0.0003 & model-implied \\
$do(A\beta\ \mathrm{low})$ & mid rate shift & -0.0082 & largest A$\beta$ response \\
$do(A\beta\ \mathrm{low})$ & late rate shift & 0.0001 & weak response \\
$do(A\beta,pTau217\ \mathrm{low})$ & mid rate shift & -0.0097 & stronger joint response \\
\bottomrule
\end{tabular}
\end{table}

BN-LTE supports g-formula contrasts by substituting selected parent variables
inside the fitted structural equations. Amyloid lowering produces its largest
model-implied rate reduction at mid pseudotime ($-0.0082$ SUVR/year), while
joint amyloid--pTau217 lowering gives a stronger mid-stage shift
($-0.0097$ SUVR/year). As observational triangulation, root-adjusted AIPW
associates amyloid positivity with faster future tau accumulation
(ATE 0.0135 SUVR/year; 95\% CI 0.0066--0.0203; ESS 527), remains positive after
propensity trimming, and is attenuated under expanded adjustment, consistent
with adjustment for variables that may track or mediate the amyloid-to-tau
pathway. Stage-stratified AIPW reinforces the mid-stage pattern. The
mid-pseudotime stratum shows the strongest observational amyloid association
(ATE 0.0366 SUVR/year; 95\% CI 0.0051--0.0681), aligning with the BN-LTE
intervention window. In contrast, the late stratum contains 11 treated subjects and only 1 control
subject, placing it outside the overlap regime needed for credible
stage-specific AIPW inference.

\subsection{Stage-Stratified Robustness}

\begin{table}[t]
\centering
\caption{Stage-stratified held-out ADNI performance}
\label{tab:stage_stratified}
\scriptsize
\setlength{\tabcolsep}{5.0pt}
\renewcommand{\arraystretch}{1.12}
\begin{tabular}{lccccc}
\toprule
Stage & Pairs & A$\beta$+ & MAE$\downarrow$ & $\rho_{\Delta}\uparrow$ & Cosine$\uparrow$ \\
\midrule
Early & 100 & 0.140 & 0.0083 & 0.670 & 0.786 \\
Middle & 40 & 0.750 & 0.0151 & 0.846 & 0.935 \\
Late & 12 & 0.917 & 0.0389 & 0.316 & 0.148 \\
\bottomrule
\end{tabular}
\end{table}

Table~\ref{tab:stage_stratified} shows that BN-LTE is most reliable in the
early-to-middle disease range, where sample size and overlap are sufficient for
interpretable subgroup analysis. Progression agreement increases from early to
middle pseudotime, reaching $\rho_\Delta=0.846$ and cosine $=0.935$ in the
middle stratum. This is also the stage where the intervention diagnostics show
the strongest amyloid-linked response, supporting a coherent mid-stage
progression signal across prediction and mechanism. The late stratum contains
only 12 held-out pairs and has limited treatment overlap, so its estimates
mainly indicate the boundary of the current data support. We therefore use the
late-stage results to motivate the need for larger late-disease cohorts, rather
than to draw definitive conclusions about late-stage BN-LTE dynamics.

\subsection{Calibration and Regional Dependence}
Table~\ref{tab:calibration_summary} shows that uncertainty calibration changes
interval reliability without altering BN-LTE point predictions. MAE and
$\rho_\Delta$ remain identical across interval models, while coverage varies
with the assumed residual structure. Nodewise diagonal and low-rank residual
intervals reach the target 90\% coverage, whereas validation-scaled and
Braak-block intervals trade narrower width for slight undercoverage. Thus,
regional dependence affects uncertainty quantification, but not the fitted
progression signal.

\begin{table}[t]
\centering
\caption{Predictive calibration and residual-dependence diagnostics}
\label{tab:calibration_summary}
\scriptsize
\setlength{\tabcolsep}{3.8pt}
\renewcommand{\arraystretch}{1.10}
\begin{tabular}{lcccc}
\toprule
Interval model & 90\% cov.$\uparrow$ & Width & MAE & $\rho_\Delta$ \\
\midrule
Nodewise diagonal & \textbf{0.923} & 0.284 & 0.0054 & 0.896 \\
Validation-scaled diagonal & 0.890 & 0.241 & 0.0054 & 0.896 \\
Braak-block variance & 0.890 & 0.241 & 0.0054 & 0.896 \\
Low-rank residual, $K=1$ & \textbf{0.923} & 0.284 & 0.0054 & 0.896 \\
Low-rank residual, $K=5$ & \textbf{0.923} & 0.284 & 0.0054 & 0.896 \\
\bottomrule
\end{tabular}
\end{table}

\section{Conclusion}
\label{sec:conclusion}

We introduced BN-LTE, a Bayesian structural framework designed to link spatial forecasting with mechanistic staging in AD progression. Across repeated ADNI experiments, BN-LTE achieved strong spatial reconstruction of regional tau-PET progression among the included baselines. More importantly, its posterior structural analyses revealed disease-time-varying AT(N) relationships that prediction metrics alone cannot capture. Future work will extend BN-LTE beyond the present ADNI-centered evaluation by applying the same structural framework to larger harmonized longitudinal cohorts such as BioFINDER-2\footnote{The Swedish BioFINDER Study: \url{https://biofinder.se/two/}}. Broader cohort coverage and more balanced sampling across disease stages will allow latent disease time to be represented as a richer posterior staging variable, support more stable estimation of AT(N)-admissible parent structure, and bring cross-region residual dependence directly into the likelihood.

\bibliographystyle{IEEEtran}
\bibliography{reference}

@misc{Jack_2018,
  title={NIA-AA research framework: toward a biological definition of Alzheimer’s disease. Alzheimers Dement. 2018; 14 (4): 535--62},
  author={Jack Jr, CR and Bennett, DA and Blennow, K and Carrillo, MC and Dunn, B and Haeberlein, SB and Holtzman, DM and Jagust, W and Jessen, F and Karlawish, J and others},
  year={2018}


}

@article{Braak_1991,
  title={Neuropathological stageing of Alzheimer-related changes},
  author={Braak, Heiko and Braak, Eva},
  journal={Acta neuropathologica},
  volume={82},
  number={4},
  pages={239--259},
  year={1991},
  publisher={Springer}
}

@article{Raj_2012,
  title={A network diffusion model of disease progression in dementia},
  author={Raj, Ashish and Kuceyeski, Amy and Weiner, Michael},
  journal={Neuron},
  volume={73},
  number={6},
  pages={1204--1215},
  year={2012},
  publisher={Elsevier}
}

@article{Raj_2015,
  title={Network diffusion model of progression predicts longitudinal patterns of atrophy and metabolism in Alzheimer’s disease},
  author={Raj, Ashish and LoCastro, Eve and Kuceyeski, Amy and Tosun, Duygu and Relkin, Norman and Weiner, Michael},
  journal={Cell reports},
  volume={10},
  number={3},
  pages={359--369},
  year={2015},
  publisher={Elsevier}
}

@article{Iturria_Medina_2014,
  title={Epidemic spreading model to characterize misfolded proteins propagation in aging and associated neurodegenerative disorders},
  author={Iturria-Medina, Yasser and Sotero, Roberto C and Toussaint, Paule J and Evans, Alan C and Alzheimer's Disease Neuroimaging Initiative},
  journal={PLoS computational biology},
  volume={10},
  number={11},
  pages={e1003956},
  year={2014},
  publisher={Public Library of Science San Francisco, USA}
}

@article{Schafer_2021,
  title={Bayesian physics-based modeling of tau propagation in Alzheimer's disease},
  author={Sch{\"a}fer, Amelie and Peirlinck, Mathias and Linka, Kevin and Kuhl, Ellen and Alzheimer's Disease Neuroimaging Initiative (ADNI)},
  journal={Frontiers in physiology},
  volume={12},
  pages={702975},
  year={2021},
  publisher={Frontiers Media SA}
}

@article{Fonteijn_2012,
  title={An event-based model for disease progression and its application in familial Alzheimer's disease and Huntington's disease},
  author={Fonteijn, Hubert M and Modat, Marc and Clarkson, Matthew J and Barnes, Josephine and Lehmann, Manja and Hobbs, Nicola Z and Scahill, Rachael I and Tabrizi, Sarah J and Ourselin, Sebastien and Fox, Nick C and others},
  journal={NeuroImage},
  volume={60},
  number={3},
  pages={1880--1889},
  year={2012},
  publisher={Elsevier}
}

@article{Giorgio_2022,
  title={A robust and interpretable machine learning approach using multimodal biological data to predict future pathological tau accumulation},
  author={Giorgio, Joseph and Jagust, William J and Baker, Suzanne and Landau, Susan M and Tino, Peter and Kourtzi, Zoe and Alzheimer’s Disease Neuroimaging Initiative},
  journal={Nature communications},
  volume={13},
  number={1},
  pages={1887},
  year={2022},
  publisher={Nature Publishing Group UK London}
}

@article{Rathore_2024,
  title={Predicting regional tau accumulation with machine learning-based tau-PET and advanced radiomics},
  author={Rathore, Saima and Higgins, Ixavier A and Wang, Jian and Kennedy, Ian A and Iaccarino, Leonardo and Burnham, Samantha C and Pontecorvo, Michael J and Shcherbinin, Sergey},
  journal={Alzheimer's \& Dementia: Translational Research \& Clinical Interventions},
  volume={10},
  number={4},
  pages={e70005},
  year={2024},
  publisher={Wiley Online Library}
}

@article{Karlsson_2025,
  title={Machine learning prediction of tau-PET in Alzheimer's disease using plasma, MRI, and clinical data},
  author={Karlsson, Linda and Vogel, Jacob and Arvidsson, Ida and {\AA}str{\"o}m, Kalle and Strandberg, Olof and Seidlitz, Jakob and Bethlehem, Richard AI and Stomrud, Erik and Ossenkoppele, Rik and Ashton, Nicholas J and others},
  journal={Alzheimer's \& Dementia},
  volume={21},
  number={2},
  pages={e14600},
  year={2025},
  publisher={Wiley Online Library}
}

@article{Jasodanand_2025,
  title={AI-driven fusion of multimodal data for Alzheimer’s disease biomarker assessment},
  author={Jasodanand, Varuna H and Kowshik, Sahana S and Puducheri, Shreyas and Romano, Michael F and Xu, Lingyi and Au, Rhoda and Kolachalama, Vijaya B},
  journal={Nature Communications},
  volume={16},
  number={1},
  pages={7407},
  year={2025},
  publisher={Nature Publishing Group UK London}
}

@inproceedings{Ozdemir_2024,
  title={A Dynamic Model for Early Prediction of Alzheimer’s Disease by Leveraging Graph Convolutional Networks and Tensor Algebra},
  author={Ozdemir, Cagri and Al Olaimat, Mohammad and Bozdag, Serdar and Alzheimer’s Disease Neuroimaging Initiative and others},
  booktitle={Pacific Symposium on Biocomputing. Pacific Symposium on Biocomputing},
  volume={30},
  pages={675},
  year={2025}
}

@article{Tekkesinoglu_2024,
  title={Explaining graph convolutional network predictions for clinicians—An explainable AI approach to Alzheimer's disease classification},
  author={Tekkesinoglu, Sule and Pudas, Sara},
  journal={Frontiers in Artificial Intelligence},
  volume={6},
  pages={1334613},
  year={2024},
  publisher={Frontiers Media SA}
}

@article{Lauber_2024,
  title={Global amyloid burden enhances network efficiency of tau propagation in the brain},
  author={Lauber, Meagan V and Bellitti, Matteo and Kapadia, Krish and Jasodanand, Varuna H and Au, Rhoda and Kolachalama, Vijaya B},
  journal={Journal of Alzheimer’s Disease},
  volume={108},
  number={1\_suppl},
  pages={S71--S80},
  year={2025},
  publisher={SAGE Publications Sage UK: London, England}
}

@inproceedings{Ren_2026_HiBrain,
  title = {HiBrain: Hierarchical Prototype Learning on Multimodal Brain Graphs for Stage-Aware Biomarker Discovery},
  author = {Ren, Jing and Yang, Kefan and Le, Nguyen Linh Dan and Zhou, Jingjing and Zhang, Xikun and Xu, Ziqi and Kong, Xiangjie and Li, Xiaodong and Xia, Feng},
  booktitle = {Proceedings of the ACM SIGKDD International Conference on Knowledge Discovery and Data Mining (KDD '26)},
  year = {2026},
  address = {Jeju, Korea},
  numpages = {12},
  url = {https://openreview.net/forum?id=elEox7tlQR},}

@article{Tabarestani_2020,
  title={A distributed multitask multimodal approach for the prediction of Alzheimer’s disease in a longitudinal study},
  author={Tabarestani, Solale and Aghili, Maryamossadat and Eslami, Mohammad and Cabrerizo, Mercedes and Barreto, Armando and Rishe, Naphtali and Curiel, Rosie E and Loewenstein, David and Duara, Ranjan and Adjouadi, Malek},
  journal={NeuroImage},
  volume={206},
  pages={116317},
  year={2020},
  publisher={Elsevier}
}

@article{Nguyen_2020,
  title={Predicting Alzheimer's disease progression using deep recurrent neural networks},
  author={Nguyen, Minh and He, Tong and An, Lijun and Alexander, Daniel C and Feng, Jiashi and Yeo, BT Thomas and Alzheimer's Disease Neuroimaging Initiative and others},
  journal={NeuroImage},
  volume={222},
  pages={117203},
  year={2020},
  publisher={Elsevier}
}

\end{document}